# The Entropy of Artificial Intelligence and a Case Study of AlphaZero from Shannon's Perspective

Bo Zhang[1*], Bin Chen[2], Jin-lin Peng[1]

**Affiliations:**

[1]Artificial Intelligence Research Center, National Innovation Institute of Defense Technology, Beijing, P. R. China.

[2]Academy of Military Sciences, Beijing, P. R. China

[*]Corresponding Author. Email: bo.zhang.airc@outlook.com

**Abstract:** The recently released AlphaZero algorithm achieves superhuman performance in the games of chess, shogi and Go, which raises two open questions. Firstly, as there is a finite number of possibilities in the game, is there a quantifiable intelligence measurement for evaluating intelligent systems, e.g. AlphaZero? Secondly, AlphaZero introduces sophisticated reinforcement learning and self-play to efficiently encode the possible states, is there a simple information-theoretic model to represent the learning process and offer more insights in fostering strong AI systems?

This paper explores the above two questions by proposing a simple variance of Shannon's communication model, the concept of **intelligence entropy** and the Unified Intelligence-Communication Model is proposed, which provide an information-theoretic metric for investigating the intelligence level and also provide an bound for intelligent agents in the form of Shannon's capacity, namely, the **intelligence capacity**. This paper then applies the concept and model to AlphaZero as a case study and explains the learning process of intelligent agent as turbo-like **iterative decoding**, so that the learning performance of AlphaZero may be quantitatively evaluated. Finally, conclusions are provided along with theoretical and practical remarks.

# Introduction

This is a previewed version of treatise. A brief review of the related research would be given in detail of a later version for journal publication. Let us directly get through the jungle of the AI's history and referred the latest success of DeepMind's AlphaZero to the recent report in Science (1), and use Shannon's communication model (2) to analyze the intelligence capacity and the learning process of intelligent agents, and provide a case study of AlphaZero.

# Unified Intelligence-Communication Model

In Figure 1, we apply our universal intelligence-communication model (UICM) proposed in (3) to AlphaZero. Specifically, there are two agents in AlphaZero to enable self-play, and they interact with each other via the environment, e.g. a 19*19 chessboard. Each agent observes the move of its opponent, evaluates the situation of the chessboard, recognizes the pattern and predicts the future actions, before it makes a decision and takes the next move. The information exchanging and processing flow is equivalent to a two-way reciprocal Shannon's communication model between agent A and B, where the communication channel is the chessboard.

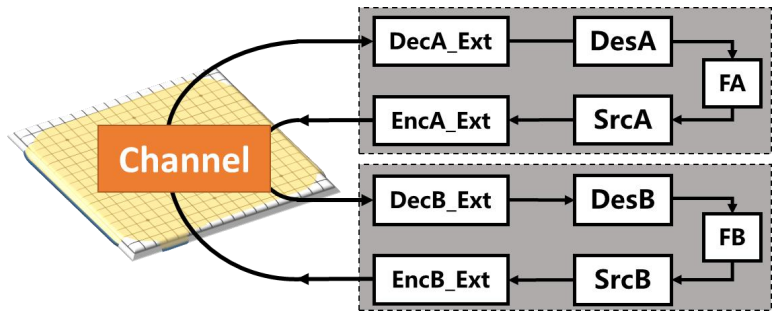

**Fig. 1: The universal intelligence-communication model for AlphaZero.**
The perception and action of two self-playing AlphaZero agents may be modelled as decoders and encoders, which capture the interaction between agents and the environment.

In chess or Go, both agents try to win the game so that each agent tries to predict the behavior of each other. Therefore, we may generalize Shannon's communication model by adding internal communication channels, which is depicted in Figure 2.

In agent A, it builds an internal environment model, including representations of the chessboard, the agent B and a critic (which is not illustrated in the figure) for evaluating the probability of winning. Therefore, agent A may play the game within itself with a virtual agent B over a virtual chessboard. This internal thinking process may also be modelled as a two-way reciprocal Shannon's communication model. In order to distinguish between the different channels, we denote the communications between the real agent A and B as *external communications*, and the communication within agents itself as *internal communications*.

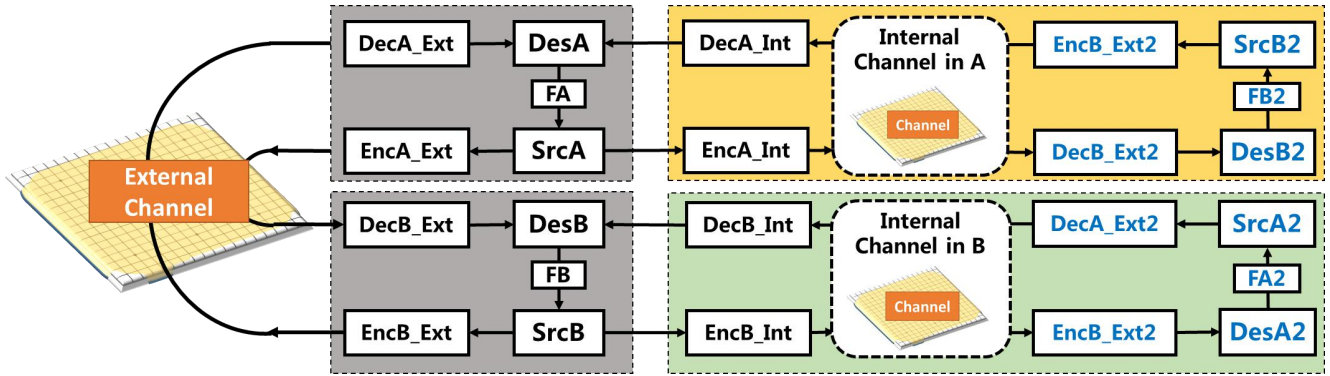

**Fig. 2: The UICM for AlphaZero with internal world models and internal channels.**
Each AlphaZero agent builds an internal channel or world model, where it virtually plays with the model of its opponent agent, predicts the effects of actions taken and learns the behavior of its opponent agent.

## Intelligence Entropy and Intelligence Capacity

Now we may formalize the goal of a single agent in AlphaZero: In the two-player zero-sum game over a communication channel, the amount of source information of agent B decoded by agent A is denoted as $I_{B-A}$, and that of agent A decoded by agent B is denoted as $I_{A-B}$. The condition of agent A dominates is $I_{B-A} > I_{A-B}$, namely, agent A conquers agent B in terms of being more certain of its opponent's strategies so that more effective actions may be taken.

Motivated by the analysis for emergence of human-level intelligence like AlphaZero, we propose the concept of Intelligence Entropy. **The Intelligence Entropy is the amount of information recovered by the agent from the environment (e.g. the external communication channel), which may be quantified by entropy.**

Therefore, the maximum amount of information that can be decoded by agent A may be quantified by the entropy of information source, which in turn, would be upper-bounded by the well-defined Shannon capacity of the external communication channel. In the case of Go, the channel capacity may be roughly quantified by the *(361!)* possible states of the chessboard, where the inequality stands for effects that the rules of Go may prohibit some of the actions taken.

$$\text{MAX}(I_{B\text{-}A}, I_{A\text{-}B}) \leq C \leq \log_2(361!) \approx 2552.$$

Here, we propose the definition of Intelligence Capacity. **Given a certain external environment and a specific task, the Intelligence Capacity denotes the maximum amount of entropy that can be extracted from the environment by the agent. It is an indicator for the highest level of intelligence for executing a specific task in a given environment, hence may define the ultimate intelligence.**

## AlphaZero Self-Play Models as an Iterative Decoder

As AlphaZero has a finite **Intelligence Capacity**, we may now take a closer look at how AlphaZero approaches the capacity by designing a sophisticated decoder.

Both competing agents are co-evolving in the internal communication channel of AlphaZero, and each agent decodes the information from the external and internal communication channel independently and iteratively. We may formulate each agent as an component decoder, as a variance of the famous turbo decoder in the information-theoretic society (3), which made a breakthrough in approaching the Shannon capacity for error-correction code design. The decoder structure can be directly extracted from the architecture of Figure 2, but we re-plot it in Figure 3 to make it more explicit.

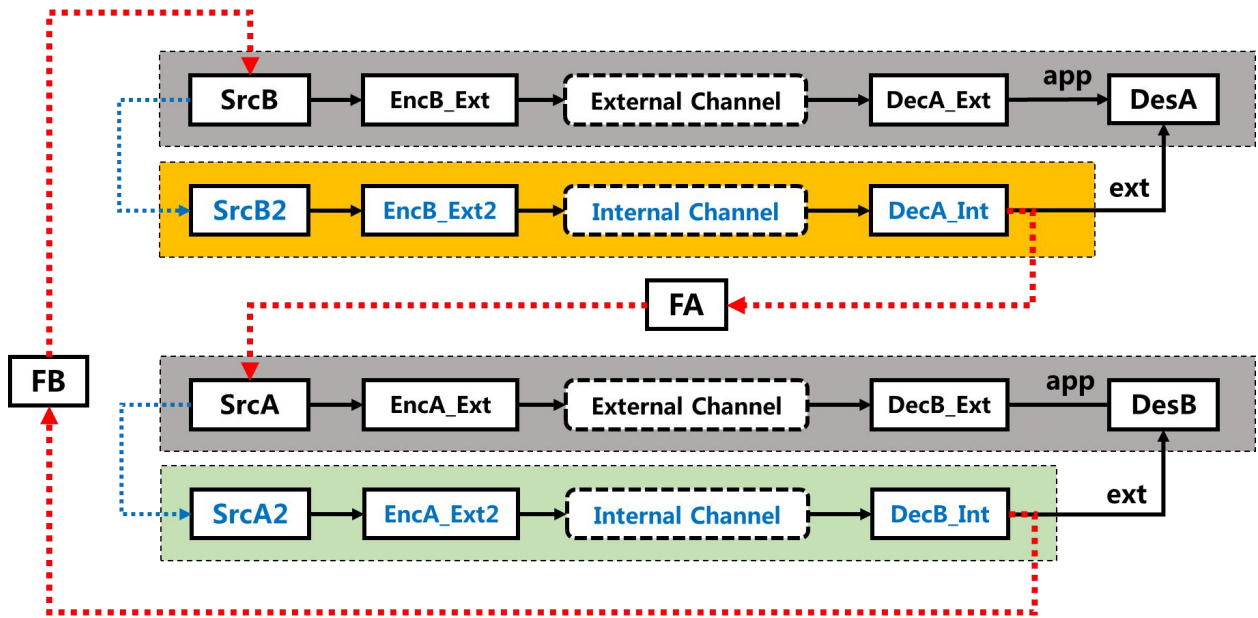

**Fig. 3: The Iterative Decoder in AlphaZero.**

Each AlphaZero agent forms a decoder for extracting information about its opponent, from a pair of external and the internal channels, which outputs extrinsic information for removing uncertainty of its opponent agent.

The main difference between a conventional turbo decoder and the proposed turbo decoder is the information source. The twin-component decoders in an interactive turbo decoder attempt to extract information from a single source. For example, the objective of agent A is to decode the information source from agent B conveyed over the external noise-free chessboard channel, so that it may decide the right move and win the game. However, as agent A cannot directly hack into the thinking pattern of agent B, a model for agent B is built within agent A itself.

Therefore, the source information of agent B rebuilt within agent A is an approximation, which improves during the learning process as well, the conversion from SrcA to SrcA2 along with the conversion in EncA_Ext2 may be integrated, forming a single encoder that may evolve over time. Also, the feedback design of FA and FB may be designed to be fully reciprocal. Hence, the structure of the iterative decoder in AlphaZero would be equivalent to standard turbo decoder (3).

## Quantitative Analysis of the Learning Process

Before delving into the quantitative analysis, an important insights is given as follows. Even though the self-playing agents are competing in terms of reducing uncertainty of each other, ***they work together to jointly decode the information over the external channel and aims at achieving the channel capacity.***

Here, we take a look at the Elo-rating metric used during the learning process of AlphaZero, where e( • ) denotes the Elo-ratings and a higher rating increases the probability of winning.

$$\Pr(A\ defeats\ B) = \frac{1}{1 + 10^{c_{elo}[e(B) - e(A)]}}$$

where e(A) or e(B) may not have an upper-bound, as long as the self-play agents matched well so that e(A)=e(B), the two component agents in AlphaZero are of equal probability of winning or losing.

Therefore, we switch our viewpoint to use Shannon's information entropy for measuring the learning process. Firstly, the intelligence capacity of AlphaZero is also an upper-bound for the intelligence capacity of the self-playing agent A or B. Secondly, as seen in Fig. 3, if the extrinsic information exchanged between the component decoders formed by the self-playing agents no longer increases, the learning process also stopped and the intelligence-level of Agent A or B stopped improving. Please note that the extrinsic information **$I_E(A)$** and **$I_E(B)$** does not necessary to reach **1.0**, as the learning process may stop at a local optimum.

This extrinsic information exchanging process may be quantitatively analyzed and graphically presented in EXtrinsic Information Transfer chart (EXIT chart) developed for analyzing decoding performance of iterative decoders (5). A successful and a unsuccessful learning process may then be distinguished by the curves in the EXIT charts, and two illustrative examples are provided in Figure 4.

*Further results for the case study in terms of the $I_E(A)$ and $I_E(B)$ curves would be provided in the version for publication.*

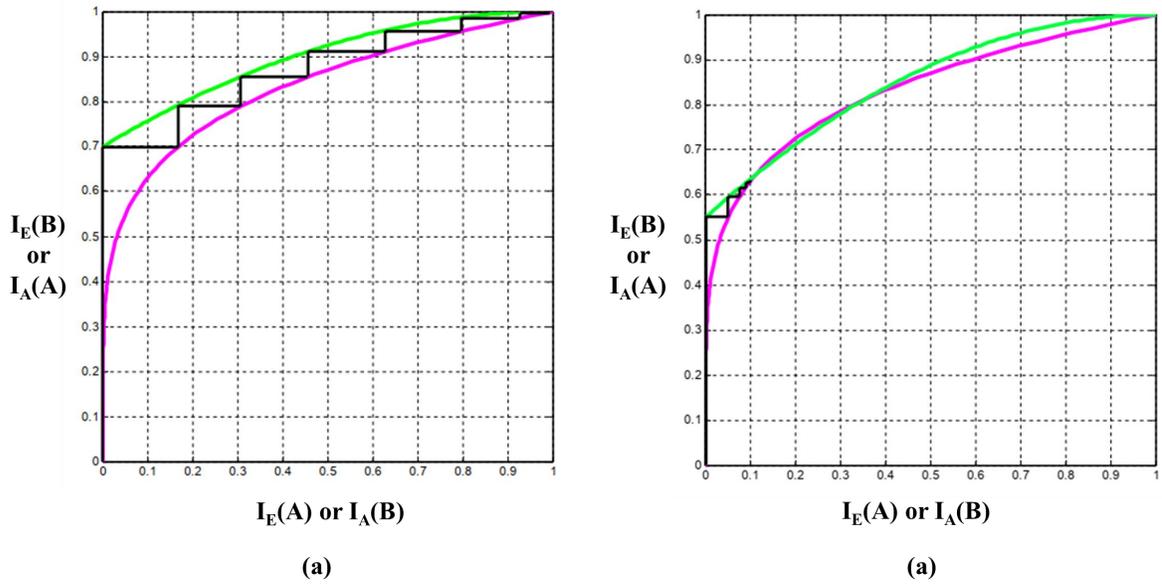

**Fig. 4: Examples of Extrinsic Information Curves.**

If the curves of extrinsic information at the two component decoders form an open tunnel from (0,0) to (1,1), the learning process is likely to be successful, depending on the model for internal channels. If the two curves intersects other than (0,0) and (1,1), the learning process generally cannot reach a global optimum.

## Conclusions and Future Works

In this paper, we modelled the interactions between an intelligent agent and its environment as a series of external and internal channels, where the intelligence entropy is proposed to measurement the intelligence of an agent and an intelligence capacity may be given in the form of the Shannon's capacity. The intelligence capacity approaching agent design was discussed, with a focus on iterative turbo-like decoder design in AlphaZero. The EXIT-chart is hence a quantitative tool for evaluating and predicting the learning process of an intelligence-bound approaching agent.

Some insights would be discussed in more detail in (3), and we briefly summarize the insights provided by AlphaZero as below:

The self-playing agents jointly decodes the information over the external channel (the world) iteratively, where the maximum amount of intelligence entropy is upper-bound by the intelligence capacity, or the external communication channel capacity. The self-playing process may been seen as an iterative decoding process, where both agents are learning to adapt to the channel (environment), to build an internal channel model (world model) that operates arbitrary closed to the external model within itself, namely, capturing the (361!) possibilities in its internal channel (the world model).

As a theoretical remark, we may define an ultimate Go player as an agent, which has the capability of achieving the intelligence capacity. If two ultimate Go players compete with each other, they have full knowledge of each other. In this case, the probability of A wins equals 50% and therefore form a quantum superposition. The uncertainty would be reduced to 0 immediately after a first move of any agent is taken and the measurement of this move leads to a collapse from the quantum superposition to a certain end.

As a practical remark, motivated by the success of AlphaZero and the analysis provided in this paper, the iterative decoding or learning philosophy may be applied to other intelligent agent designs, and the internal channel may be modelled by component structure such as deep neural networks to approximate the external channel (world model). Hence, intelligence capacity approaching learning systems may be built following an iterative decoder architecture, which may bring about a breakthrough of performance in comparison to the state-of-art single component decoder design.

EXIT charts may serve as a powerful tool for evaluating the design of intelligence capacity approaching learning systems, meanwhile the tracking of the mutual information becomes an important issue in dealing with the situation that the agent may execute a variety of tasks in a dynamic and open environment, as it is in general of much more uncertainty than AlphaZero, where the agent execute a well-defined task of Go in a static and closed environment of chessboard.

Finally, we are pursuing the quantitative theories and methods for building human-level strong AI that may adapt to changing environment and tasks. The intelligence entropy and capacity may become useful tools to explain how does human intelligence emerge from the biological wet-ware of 100 billion neurons connected by 100 trillion synapses.